\documentclass[letterpaper, 10 pt, conference]{ieeeconf}  

\IEEEoverridecommandlockouts                              

\title{\LARGE \bf
Dexterous In-hand Manipulation by Guiding Exploration \\ with  Simple Sub-skill Controllers
}

\author{Gagan Khandate\authorrefmark{1}\authorrefmark{2}, Cameron Paul Mehlman\authorrefmark{1}\authorrefmark{3}, Xingsheng Wei\authorrefmark{1}\authorrefmark{3}, Matei Ciocarlie\authorrefmark{3}
\thanks{\authorrefmark{1} denotes joint first authorship}%
\thanks{\authorrefmark{2}Dept. of Computer Science, \authorrefmark{3}Dept. of Mechanical Engineering}%
\thanks{Columbia University, New York, NY 10027, USA}%
\thanks{Corresponding email:~\texttt{gagank@cs.columbia.edu}}%
\thanks{This work was supported in part by
the ONR under grant N00014-21-1-4010 and the NSF under grant CMMI-2037101.}
}

\usepackage[sorting=none, maxcitenames=1, maxbibnames=99, style=numeric-comp]{biblatex}
\AtBeginBibliography{\small}
\usepackage{graphicx}
\usepackage[font=small]{caption}
\usepackage{amsmath}
\usepackage{amssymb}
\usepackage{xcolor}
\usepackage{hyperref}
\hypersetup{colorlinks,linkcolor={black},citecolor={green},urlcolor={gray}}

\usepackage{algorithm} 
\usepackage[noend]{algpseudocode} 
\usepackage{lipsum}
\usepackage{subcaption}

\usepackage{booktabs}
\newcommand{\ra}[1]{\renewcommand{\arraystretch}{#1}}

\newcommand{\bm}[1]{\boldsymbol{#1}}

\begin{document}
\maketitle
\thispagestyle{empty}
\pagestyle{empty}

\begin{abstract}
Recently, reinforcement learning has led to dexterous manipulation skills of increasing complexity. Nonetheless, learning these skills in simulation still exhibits poor sample-efficiency which stems from the fact these skills are learned from scratch without the benefit of any domain expertise. In this work, we aim to improve the sample efficiency of learning dexterous in-hand manipulation skills using controllers available via domain knowledge. To this end, we design simple sub-skill controllers and demonstrate improved sample efficiency using a framework that guides exploration toward relevant state space by following actions from these controllers. We are the first to demonstrate learning hard-to-explore finger-gaiting in-hand manipulation skills without the use of an exploratory reset distribution. 

\end{abstract}

\section{Introduction}



Reinforcement learning has led to significant improvements in our ability to achieve a range of dexterous manipulation skills. Prior works have demonstrated in-hand manipulation skills such as object reorientation to a desired pose \cite{OpenAI2018-bx, Chen2022-ud} or continuous re-orientation about a given axis \cite{Khandate2022-qt, Qi2022-wy}. However, learning these dexterous skills can still require up to billions of simulation steps.


Methods using domain expertise to improve the sample efficiency for learning in-hand manipulation skills achieve this mainly through engineered reset distributions and model-based controllers. The methods using reset distributions use a set of stable grasps to enable exploration which has been particularly effective for finger-gaiting in-hand manipulation tasks\cite{Khandate2022-qt}. However, the reset distributions are object and task specific and may require heavy engineering. 

Alternatively, other methods use model-based controllers that leverage the model to ensure the stability of the grasp and combine them with reinforcement learning. Predominantly, this involves decomposing the in-hand manipulation task into appropriate sub-skills (i.e in-grasp manipulation, finger-pivoting, finger-gaiting, etc,.) and designing model-based controllers for each to be used as low-level controllers in hierarchical reinforcement learning \cite{Li2019-fq}.

However, this approach of using model-based controllers as low-level controllers has a number of limitations. First, from a training perspective, computing actions from these controllers can be expensive, particularly if it involves optimization. Hence, querying these controllers at every step can outweigh the benefits in terms of the net computational cost. Next, restricting the policy to only executing one of the predefined sub-skills can lead to learning a sub-optimal policy. Finally, a hierarchical learning framework implicitly requires that low-level model-based controllers must be deployable on hardware. Thus, these model-based controllers must be designed such that only sensor feedback available on the hardware is used as input, and controllers must be robust to uncertainties present on the hardware.

\begin{figure}[t]
\centering
\includegraphics[width=0.45\textwidth]{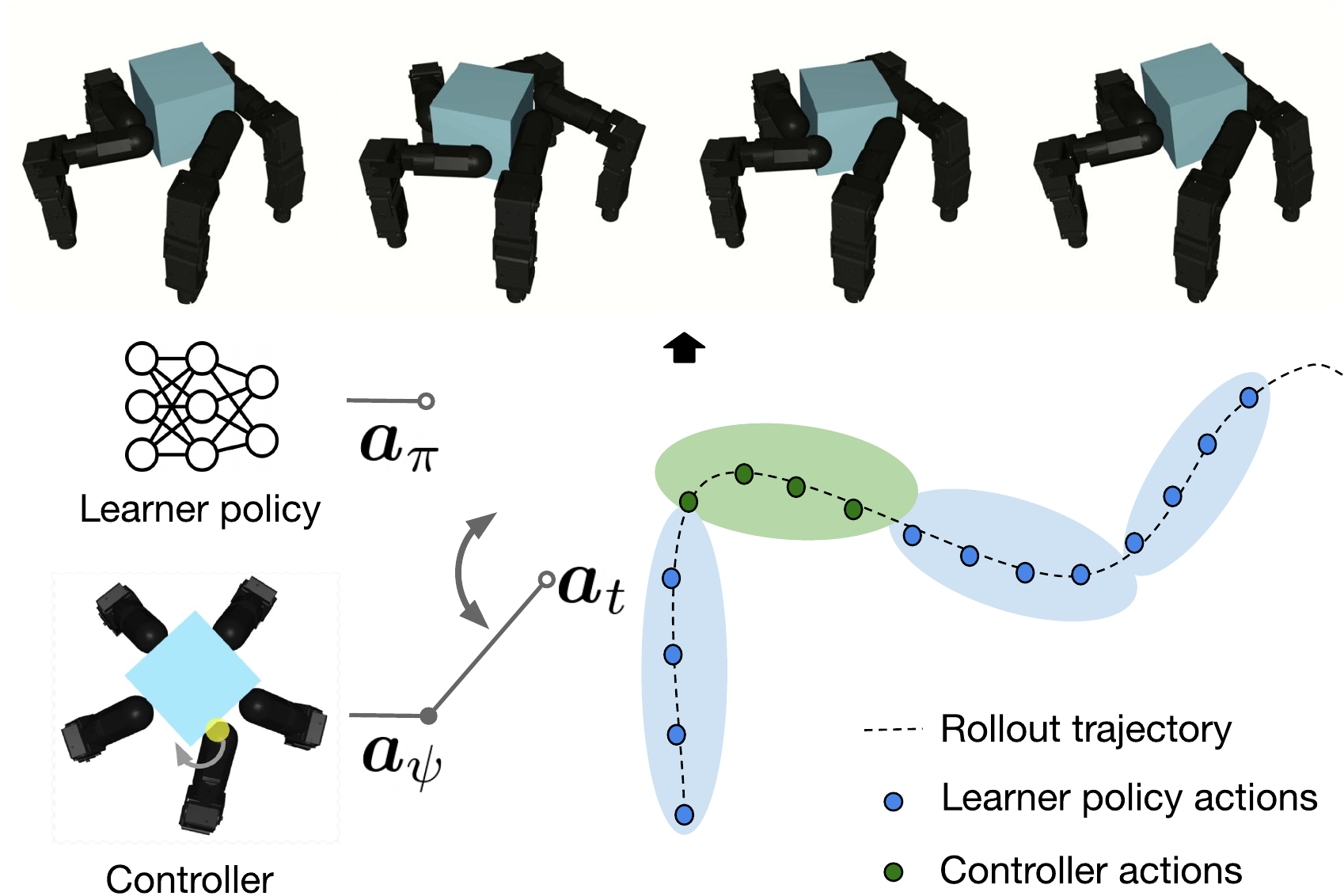}
\caption{Our method of interweaving the policy and sub-skill controller (ex. contact switching controller) during training allows the policy to effectively learn dexterous finger-gaiting skills as shown. Videos can be found at project page: \href{https://roamlab.github.io/vge}{roamlab.github.io/vge}}
\label{fig:eyecandy}
\end{figure}

In this work, we seek to leverage model-based controllers to improve sample efficiency while retaining the benefits of learning an end-to-end policy. The key idea that we propose is to interleave learner policies and sub-skill controllers during episode rollout during the initial phase and use only the learner policy during the later phase of training. Consequently, we use the sub-skill controllers only during training and not at run-time. We show that our method guides exploration towards relevant regions of the state-space. We also use the current action-value estimates provided by the critic to guide action selection which further improves exploration. 

Critically, we use controllers that are simple to design and inexpensive to query actions from. Our controllers do not involve multi-step horizon motion planning or other sources of complexity and computational cost. This makes them inexpensive to compute - favorably reducing the cost of querying the expert. However, inexpensive controllers will often be sub-optimal in their behavior. Still, our method enables learning despite using such sub-optimal controllers. Overall, our main contributions here include:
\begin{itemize}
    \item We demonstrate that a set of simple sub-skill controllers can be leveraged to learn dexterous in-hand manipulation tasks such as finger-gaiting.
    \item In contrast to previous work combining model-based controllers and learning for manipulation, we use the proposed controllers for training an end-to-end policy that does not require these controllers for deployment. Our approach alleviates many constraints on the controllers we use, in terms of both performance and sensory input.
    \item We also show that sub-optimal controllers can enable effective exploration of the complex state space of our problem, without exploratory reset distributions used in previous studies with similar goals. 
\end{itemize}


\section{Related Work}


Dexterous in-hand manipulation has long been of interest and numerous methods have been proposed ranging from simple low-level feedback primitive controllers to trajectory optimization for the complete task. Early model-based work on finger-gaiting \cite{Leveroni1996-iy, Han1998-xj, Saut2007-su} and finger-pivoting \cite{Omata1996-hp} propose controllers using simplified models. \textcite{Fan2017-vz, Sundaralingam2018-zw} use model-based online trajectory optimization and demonstrate finger-gaiting in simulation. \textcite{Chen2021-oo} combine trajectory optimization and tree search to follow a desired object pose trajectory.

Recently, remarkable progress in dexterous manipulation has been achieved with model-free RL, typically, demonstrating reorientation to a desired object pose \cite{OpenAI2018-bx}. However, model-free RL requires hundreds of hours of training. It was shown that GPU physics could be used to accelerate learning these skills \cite{Makoviychuk2021-ko, Allshire2021-qp}. \textcite{Chen2021-ig, Chen2022-ud} demonstrated in-hand re-orientation for a wide range of objects under palm-up and palm-down orientations of the hand with extrinsic sensing providing dense object feedback. While \textcite{Khandate2022-qt} showed object-agnostic dexterous finger-gaiting and finger-pivoting skills using precision fingertip grasps and sensing intrinsic to the hand only. \textcite{Qi2022-wy} used rapid motor adaptation to achieve effective sim-to-real transfer of in-hand manipulation skills for small cylindrical and cube-like objects. Nonetheless, learning these skills still remains sample inefficient.

To improve sample complexity, model-based controllers can be used with model-free reinforcement learning. The natural approach to leverage model-based controllers is to use them as low-level policies in hierarchical reinforcement learning. \textcite{Li2019-fq} derives model-based controllers for reposing, sliding, and flipping, and then learn a hierarchical policy. However, they demonstrate it only for 2D in-hand manipulation.  \textcite{Veiga2018-oq, Veiga2020-zm} used a low-level controller to maintain grasp stability based on tactile feedback but only evaluated for in-grasp manipulation. These approaches all share a common limitation -- the low-level controllers are necessary not only during training but also during deployment. This restricts the feedback used by low-level controllers to the sensing modalities available on the hand while simultaneously requiring the controller to be robust to the sim-to-real gap. 

Learning a single end-to-end policy while using model-based controllers only during the training phase circumvents this issue. While numerous methods suitable have been proposed, these methods possess limitations for learning dexterous manipulation with sub-optimal controllers.

Learning from demonstrations collected using the model-based controllers can be done either through augmentation \cite{Rajeswaran2017-au, Radosavovic2021-pg} or offline RL \cite{Levine2020-pe}. However, demonstrations collected using our sub-skill controllers will not be sufficiently exploratory as they can be used for only short rollout horizons. 


Imitation learning (IL) methods such as DAgger \cite{Ross2010-kp} and AggreVaTe \cite{Ross2014-qk} are effective frameworks but only for imitation with a single expert. Moreover, these methods also require the expert to be near-optimal, however, a near-optimal expert controller for dexterous manipulation is both challenging to achieve and expensive to compute. 


Adaptions modifying the behavior policy of off-policy RL with sub-skill controllers have also been considered as an alternate approach. \textcite{Kurenkov2019-ye} proposed AC-Teach framework to sample from multiple experts based on the critic's estimate of the action-value for non-prehensile manipulation. \textcite{Jeong2020-pt} derived REQ to better utilize the off-policy transitions sampled from a sub-skill expert. Here, we also use the AC-Teach framework, as it provides an effective method to incorporate sub-skill controllers for learning an end-to-end policy, and show that simple controllers and highly sub-optimal controllers can enable exploration even for highly dexterous motor control tasks such as stable in-hand manipulation.

\section{Method}
\label{sec:methods}
We will demonstrate our method for learning dexterous manipulation skills on a task exhibiting a particularly challenging problem in exploration. Specifically, we consider the task of finger-gaiting in-hand manipulation with only fingertip grasps and no support surfaces - a necessary skill for continuous object re-orientation in arbitrary orientations of the hand. Learning this skill purely with reinforcement learning is challenging as random exploration from a fixed state does not sufficiently explore the desired state-space \cite{Khandate2021-wl}. Due to the complexity and contact-rich nature of the task, it also challenging to use model-based methods to design an expert for finger-gaiting.

Finger-gaiting is a rich skill that consists of a range of behaviors that we can split into two sub-skills: in-grasp manipulation, and contact switching. In-grasp manipulation entails maintaining the object in a stable grasp and reorienting it without breaking or making contact. Therefore, due to limits on the range of motion of the hand, the maximum object rotation is also limited. Contact switching involves breaking and making contact to re-grasp the object in order to further object reorientation via in-grasp manipulation.

Thus, instead of designing a single controller for finger-gaiting we can decompose it into two controllers, an in-grasp manipulation controller and a contact switching controller. Still, ensuring optimality and robustness is challenging for these controllers. Nevertheless, sub-optimal model-based controllers obtained from simplified models or heuristics can still generate reasonable actions and are able to generate exploratory sequences of state transitions. 

In this work, we will (i) design simple controllers for in-grasp manipulation and contact switching, and (ii) learn an effective end-to-end policy for finger-gaiting using these sub-optimal controllers by following their actions.



\subsection{Problem definition and sub-optimal controllers}
\label{sec:learnfg}
We consider the task of finger-gaiting in-hand manipulation by learning continuous re-orientation about z-axis in a hand-centric frame. We aim to learn such a policy by rewarding angular velocity of the object along the z-axis. Let us consider a fully dexterous hand with $m$ fingers and $d$ dofs. Assuming position control, our policy outputs joint target updates from observations $\bm{s}$:
\begin{equation}
    \bm{s} = [\bm{q}, \bm{q}', \bm{p}_1, .. , \bm{p}_m, \bm{n}_1, .., \bm{n}_m, c_1, .., c_m]
\end{equation}
where $\bm{q} \in \mathcal{R}^d$, $\bm{q}'\in \mathcal{R}^d$ are current and target joint positions, $\bm{p}_i \in \mathcal{R}^3$, $\bm{n}_i \in \mathcal{R}^3$ are contact positions, contact normal in the global coordinate frame for contacts on all fingertips and $c_i \in \mathcal{R}$ is the magnitude of contact forces. Similar to~\cite{Khandate2022-qt}, we train this policy by rewarding the angular velocity of the object about the z-axis when making at least 3 contacts between the object and the fingertips.

As previously discussed, finger-gaiting skill can be broken down into two sub-skills - in-grasp manipulation and contact switching. We will now design controllers for each.

\begin{figure}[t]
\centering
\includegraphics[width=0.4\textwidth]{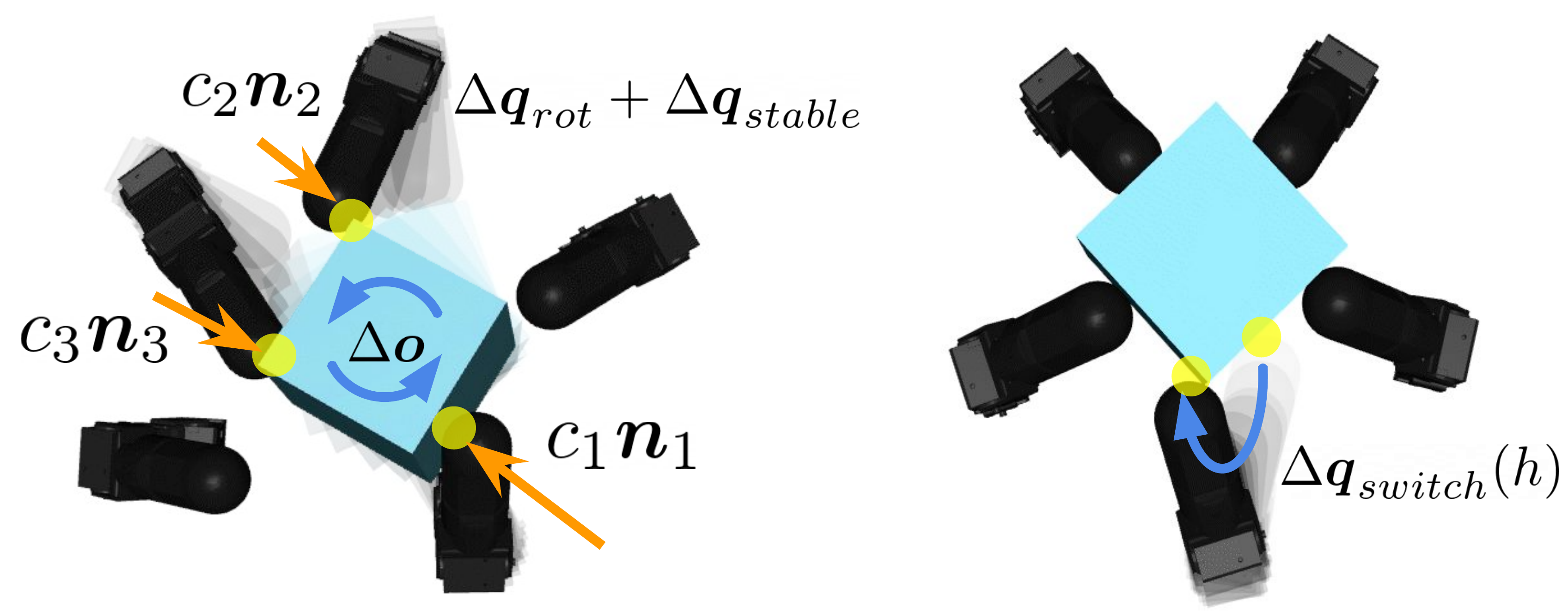}
\caption{Sub-skill controllers used for learning finger-gaiting: (left) In-grasp manipulation controller (right) Contact switching controller as described in Sec \ref{sec:learnfg}}.
\label{fig:experts}
\vspace{-5mm}
\end{figure}

\subsubsection{In-grasp manipulation controller (IGM)}
\label{sec:igm}
The in-grasp manipulation sub-skill maintains a stable grasp on the object and, simultaneously, reorients the object about the desired axis. We construct this controller by superposition of two lower-level controllers - one responsible for stability and another responsible for object re-orientation. 

To maintain grasp stability, we assume at least three contacts and desire the net object wrench be close to zero. We use the kinematic model of the hand to derive our stability controller as follows. Let $S$ be the set of contact points and $\bm{n}_1, \ldots, \bm{n}_k$ ($k\geq3$) be the contact normals. $\bm{G}_S$ and $\bm{J}_S$ are the grasp map matrix and the Jacobian for the set of contacts.  Our stability controller keeps the object in equilibrium by minimizing the net wrench applied to the object. This can be achieved by solving the following Quadratic Program (QP):
\begin{align}
\text{unknowns: normal}&\text{ force magnitudes }c_i,~i=1 \ldots k \nonumber \\
\text{minimize}~\|\bm{w}\|&~\text{subject to:} \nonumber \\
\bm{w} &= \bm{G}_S^T \left[ c_1 \bm{n_1} \ldots c_k \bm{n_k} \right]^T \label{eq:stabstart}\\
c_i & \geq  0~\forall i \\
\exists j~\text{such that}~c_j &= 1~\text{(ensure non-zero solution)}
\end{align}

Let $c_i^*$'s be the desired contact forces obtained as solution of above QP. We can compute the actions $\Delta q_{stable}$ needed for stability (expressed as joint position setpoint changes) by solving the following system: 
\begin{eqnarray}
    \Delta \bm{p}_i &=& \alpha (c^*_i - c_i)\bm{n}_i\\
    \bm{J}_S \Delta q_{stable} &=& [\Delta \bm{p}_1, \ldots, \Delta \bm{p}_k]^T 
    \label{eq:dqstab}
\end{eqnarray}
where $\alpha$ is the stability controller gain.

Separately, let $\Delta \bm{o} \in \mathcal{R}^6$  denote the desired change in object pose. We can compute the action $\Delta \bm{q}_{rot}$ required to  achieve this object re-orientation by solving
\begin{align}
    \bm{J}_S \Delta \bm{q}_{rot} = \bm{G}_S^{T} \Delta {\bm{o}}
    \label{eq:dqrot}
\end{align}

Finally, $\Delta \bm{q}_{stable}$ and $\Delta \bm{q}_{rot}$ are combined together simply via superposition. The complete set of steps involved in the in-grasp manipulation controller is outlined in Alg.~\ref{algo:ingrasp}. When used by itself in a rollout the cumulative reward achieved by the in-grasp manipulation controller before reaching joint limits is far lower than the cumulative reward achievable by a successful finger-gaiting policy within the same duration. 


\begin{algorithm}[t]
\caption{In-grasp manipulation (IGM) controller}
\begin{algorithmic}[1]
\State Get Jacobian, $\bm{J}_S$ and grasp map matrix $\bm{G}_S$
\State Compute optimal contact force $c^*_i$'s for stability from QP
\State Solve for $\Delta \bm{q}_{stable}$ using Eq~\ref{eq:dqstab}
\State Solve for $\Delta \bm{q}_{rot}$ from sampled $\Delta \bm{o}$ using Eq~\ref{eq:dqrot}
\State $\psi_{IGM}(\bm{s}) = \Delta \bm{q}_{stable} + \Delta \bm{q}_{rot}$ 
\end{algorithmic}
\label{algo:ingrasp}
\end{algorithm}

\subsubsection{Contact switching controller (CS)}
\label{sec:cs}
Our insight here is to simply use a fixed trajectory as a controller, 
\begin{equation}
    \psi_{CS}(\bm{s}) = \Delta \bm{q}_{switch}(h)  
\end{equation}
where $h=1, \ldots, H$ and $H$ is the length of the controller trajectory. 

This trajectory is a simple primitive for breaking and making one contact as described next. We design a trajectory that only breaks and makes contact with one selected finger at a given time even though contact switching as a skill can involve multiple contact switches.  We select a finger in contact at random and follow a hand-designed trajectory as shown in Fig~\ref{fig:experts} for the selected finger to break contact and remake contact with the object at a different location. This trajectory is achieved as shown in Fig~\ref{fig:switchtraj}. As one can expect, the switching controller by itself collects no meaningful reward.

\subsubsection{Finger-gaiting controller (FG)} Finally, we also construct a third controller. Combining the in-grasp manipulation and contact switching controller we construct a finger-gaiting controller. While a randomly selected fingertip breaks and makes contact, this controller also reorients the object with the fingertips in contact. Although this controller comes closer than the rest, it does not achieve the whole finger-gaiting skill we are looking to learn. Continuous object re-orientation is not possible with this controller as the fingers selected to switch contacts are still selected at random without any coordination.

We construct this controller again by superposition. We superpose in-grasp manipulation and contact switching controller.
\begin{align}
    \psi_{FG}(\bm{s}) &= \psi_{IGM}(\bm{s})  + \psi_{CS}(\bm{s}) 
\end{align}


This finger-gaiting controller also achieves a far lower return relative to a successful finger-gaiting policy.

\subsection{Learning from sub-skill controllers}


We use off-policy actor-critic reinforcement learning, where the policy used for episode rollouts (i.e the behavior policy) can be significantly different from the end-to-end policy being learned (i.e the learner policy). We use this dichotomy to enable exploration by constructing a behavior policy that is a composition of the learner policy and the sub-skill controllers. Importantly, this approach also sidesteps the need for the controller during deployment - once trained, the learner policy by itself is sufficient for deployment.

We construct the behavior policy to achieve sufficient exploration as follows. Our behavior policy periodically selects between the controllers and learner policy. We bias this selection towards higher value controllers using the critic's action-value estimate of the actions sampled from these controllers. This heuristic allows the probability of selecting a controller action to vary across the state-space. We show that this heuristic of using action-value estimates to select between available controllers and learner policy improves exploration. Optionally, we use value-weighted behavior cloning in addition to the nominal RL objective of the chosen off-policy RL method to update the policy. The method is detailed in the following paragaphs.



Let $\pi_\theta$ represent the learner policy and $\psi$ represent the controller. At every step of the rollout, we query actions from both the learner policy and controllers. Let these actions be $\bm{a}_\pi$, $\bm{a}_{\psi}$. The probability $p_{\psi}$ of selecting the action from a controller, as well as the probability $p_{\pi}$ of selecting the action prescribed by the learner policy, are computed as:


\begin{align}
p_{\psi} &= \frac{\exp(Q(\bm{s},\bm{a}_{\psi}))}{\exp(Q(\bm{s},\bm{a}_\pi)) + \exp(Q(\bm{s},\bm{a}_{\psi}))} \label{eq:softmaxpsi}\\
p_\pi &= 1 - p_{\psi} \label{eq:softmaxpi}
\end{align}

As $Q(\bm{a}_\pi, \bm{s})$ increases, $p_{\psi}$ decreases, i.e., as the learner policy improves, the probability of following the controller decays exponentially. In practice, we also impose a hard stop, i.e., we stop querying the controllers altogether after a few million steps of training and only use the learner policy for exploration. This point typically corresponds to the convergence of the critic. 

\begin{algorithm}[t]
\caption{Learning from sub-optimal controllers}
\label{algo:qsampling}
\begin{algorithmic}[1]
\Require Controller $\psi$ \\
Initialize policy $\pi_{\theta}$, $Q_{\phi}$ and off-policy RL algorithm with replay buffer $\mathcal{D}$
\For{each iteration}
\For{$t= 0, \ldots, T-1$}
\For{every $H$ steps} \label{alg:line:behavstart}
\State Compute $p_{\pi}$, $p_{\psi}$ from $\bm{a}_\pi, \bm{a}_{\psi}$ (Eq~\ref{eq:softmaxpsi} \& \ref{eq:softmaxpi})
\State Select $\mu$: $\mu \sim \{ \pi, \psi \}$ given $p_{\pi}$, $p_{\psi}$ 
\EndFor
\State Follow selected $\mu$ for next H steps: $\bm{a}_t \sim \mu(\bm{s}_t)$ \label{alg:line:behavend} 
\State $\mathcal{D} \leftarrow \mathcal{D} \cup \{\bm{s}_t, \bm{a}_t, r(\bm{s}_t, \bm{a}_t), \bm{s}_{t+1}\}$
\EndFor
\For{each gradient step}
\State $\theta \leftarrow \theta - \nabla_{\theta} \mathcal{L}_{RL}(\theta)$ 
\State $\theta \leftarrow \theta  -  \nabla_{\theta} \mathcal{L}_{BC}(\theta)$ (Optional)
\label{alg:line:bcloss}
\EndFor
\EndFor
\end{algorithmic}
\end{algorithm}

Our method is summarized in Alg~\ref{algo:qsampling}. The first difference w.r.t the standard off-policy learning framework is steps \ref{alg:line:behavstart}-\ref{alg:line:behavend} when the behavior policy is different from the learning policy and is composed as described previously. The second difference, in step \ref{alg:line:bcloss}, entails updating the learner policy parameters using a value weighted behavior cloning loss $\mathcal{L}_{BC}$ (see Eq \ref{eq:bcloss}), in addition to the nominal actor loss $\mathcal{L}_{RL}$ as determined by the off-policy algorithm of choice. We follow a linear decay schedule for $\beta$.

\begin{equation}
    \mathcal{L}_{BC}(\theta) = \sum_{(\bm{s}, \bm{a}),~\bm{a} \neq \bm{a}_\pi }\beta \log \pi_{\theta}(\bm{a} | \bm{s}) Q(\bm{s}, \bm{a})
    \label{eq:bcloss}
    \vspace{-5mm}
\end{equation}


~\\


\section{Experiments and Results}
\label{sec:results}
Our main experimental goals are: (a) illustrating the importance of sampling actions from controllers for enabling exploration and (b) studying the effect of the different sub-optimal controllers proposed.  


\begin{figure*}[]
\centering
\includegraphics[width=0.9\textwidth]{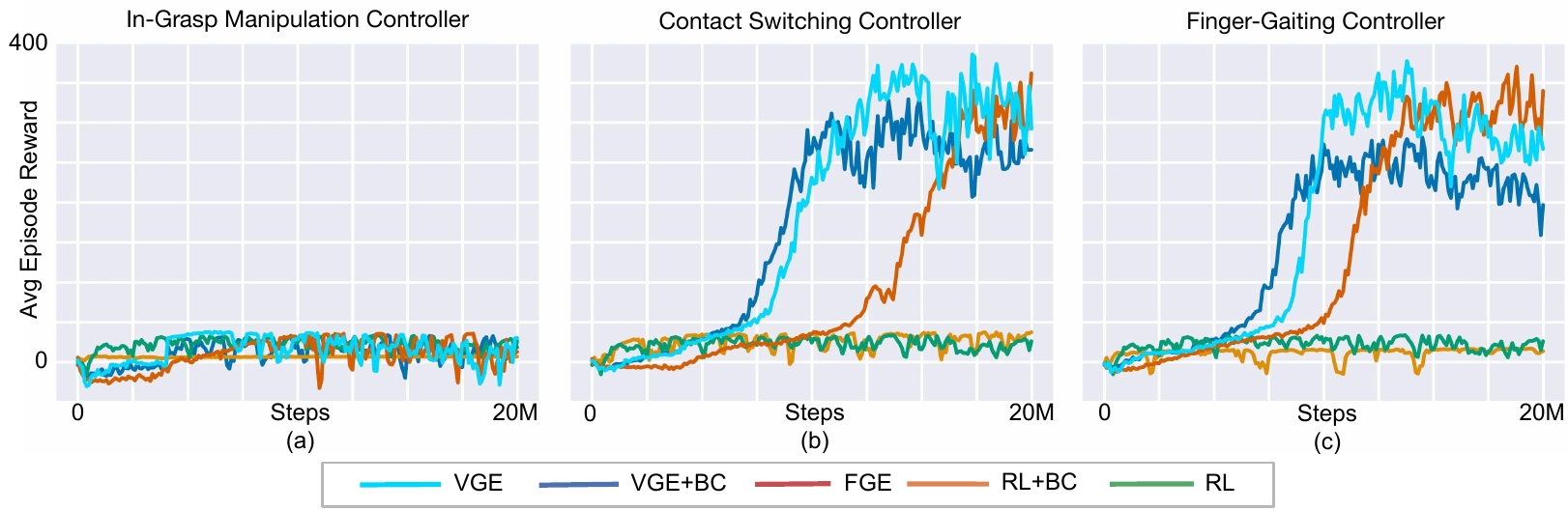}
\caption{Training curves for using (a) In-grasp manipulation controller (b) Contact switching controller (c) Finger-gaiting controller over all the evaluation conditions listed in Sec~\ref{sec:comp}. Our method that interleaves following controllers with policy (VGE, VGE + BC) learns finger-gaiting, while standard off-policy RL that does not follow controller actions fails. This shows it is essential to follow actions sampled by the sub-skill controllers to enable exploration.}
\label{fig:results}
\vspace{-5mm}
\end{figure*}

\begin{figure}
\centering
\begin{subfigure}[b]{0.13\textwidth}
\includegraphics[width=\textwidth]{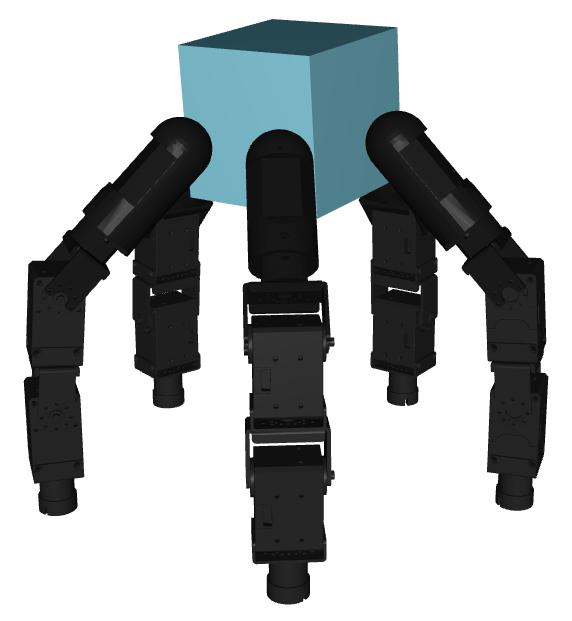}
\caption{15-dof hand}
\label{fig:hand}
\end{subfigure} 
\hfill
\begin{subfigure}[b]{0.25\textwidth}
\centering
\includegraphics[width=\textwidth]{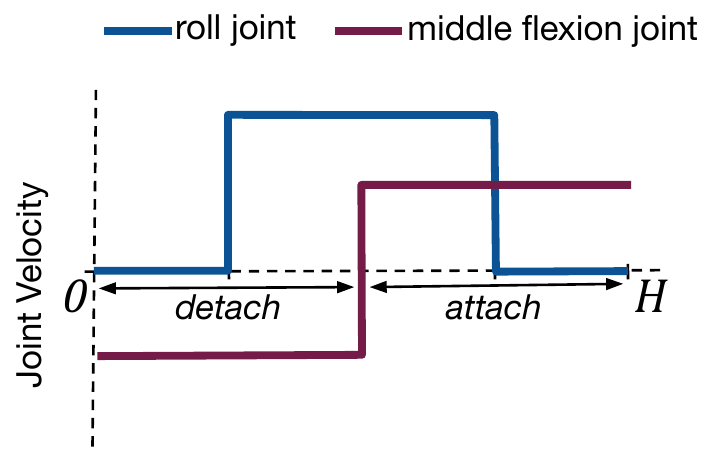}
\caption{Joint velocity for contact switching}
\label{fig:switchtraj}
\end{subfigure}
\caption{(a) The 5-fingered, 15-dof dexterous hand grasping the cube we use in our experiments (b) Joint velocity profile of the joints of the finger making and breaking contact. It involves detach and attach phases. Note that the distal flexion joint is held fixed.}
\end{figure}

\subsection{Experimental setup}
\label{sec:comp}

We use a dexterous hand in a simulated environment as shown in Fig~\ref{fig:hand}. It is a fully-actuated $15$-DOF hand consisting of five fingers where each finger consists of one roll joint followed by two flexion joints. Although our method is effective for a range of objects, in our analysis, we use a cube of side $10cm$ unless otherwise specified.  

The key hyperparameters used for the sub-optimal controller are as follows. For the in-grasp manipulation controller, we set the desired change in object pose $\Delta\bm{o}$ so that it represents a small reorientation about the z-axis. For the contact switching controller, which follows a fixed hand-designed trajectory, the velocity profile of the trajectory is as shown in Fig~\ref{fig:switchtraj}. It is designed to detach and re-attach the fingertip with the object. The roll-joint and middle-flexion-joint velocities are randomly sampled in the range of $[0.1, 0.4]$ rad/s.

Finally, as our off-policy RL algorithm, we used SAC \cite{Haarnoja2018-zj} and TD3 \cite{Fujimoto2018-lk} both with similar results. However, for brevity, we show results only for SAC. Note that with SAC, we set the exploration co-efficient $\alpha = 0$ -- using any non-zero value fails for all the baselines and also our method(s). Additionally, we do not query the controllers for actions after 4M steps of training in all of our evaluations.

\subsection{Evaluated conditions}
\label{sec:evalcond}

\begin{table}
    \centering
    \caption{Training robustness calculated as the percentage of seeds that successfully learn with about 10 seeds for each method.}
    \ra{1.1}
    \begin{tabular}{ccc}
        \midrule
         \phantom{} & CS Controller &  FG Controller\\
         \midrule
         VGE & $27\%$ &  $\mathbf{100}\%$ \\
         VGE + BC &  $36\%$ & $\mathbf{100}\%$ \\
         FGE & $67\%$  & $\mathbf{73}\%$\\
         \midrule
    \end{tabular}
    \label{tab:robust}
\end{table}

\label{sec:resultdisc}
\begin{figure}[t]
\centering
\includegraphics[width=0.35\textwidth]{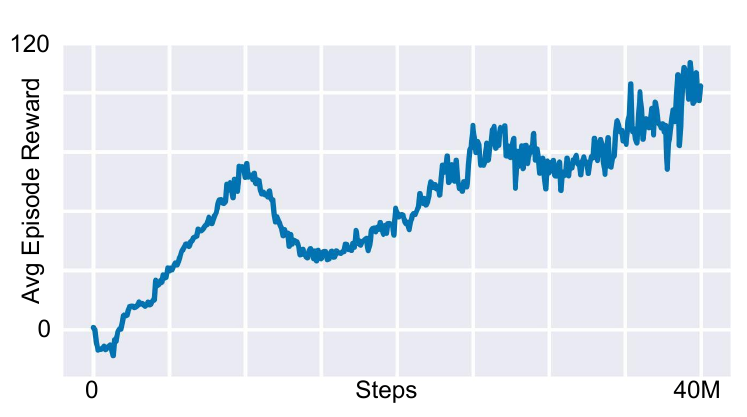}
\caption{Training curve for multiple objects using finger-gaiting controller.}
\label{fig:multiobj}
\vspace{-5mm}
\end{figure}

\begin{figure*}[]
\centering
\includegraphics[width=0.7\textwidth]{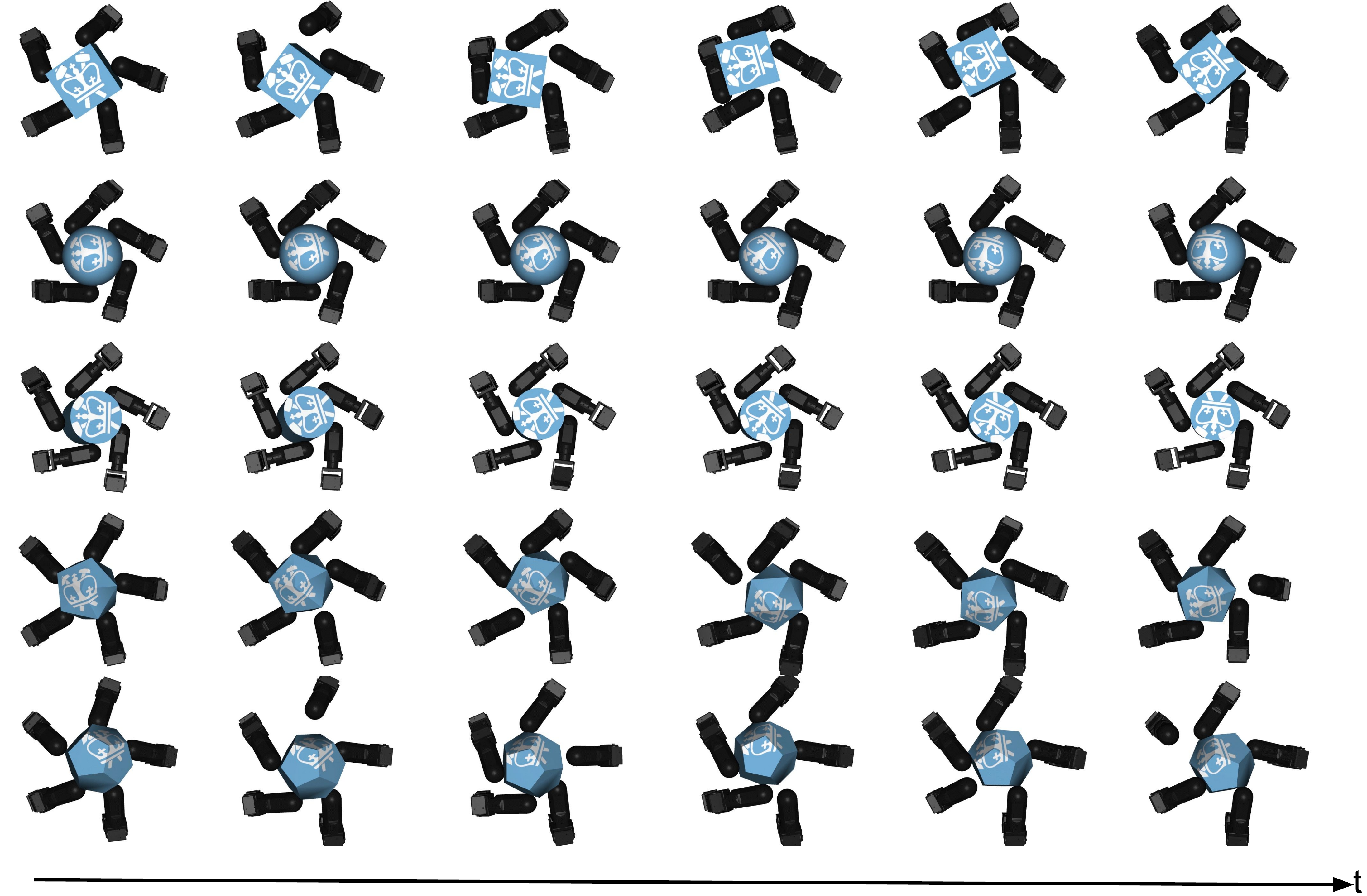}
\caption{Keyframes of the gaits achieved for multiple objects simultaneously with a single policy.}
\label{fig:gaits}
\end{figure*}

We evaluate and compare the following approaches:
 
\subsubsection{RL} Standard off-policy RL without the use of controllers and without any modification to the behavior policy or loss function is the first baseline. 

\subsubsection{RL + BC loss} We also test a version of off-policy RL augmented using the same behavior cloning loss w.r.t the behavior policy as used in our method. 




\subsubsection{Fixed Guided Exploration (FGE)} We consider this baseline to study the advantage of allowing the probability of following the controllers to vary across the state-space. Unlike our method, controller probability is independent of the state as it is a fixed decay schedule based on training progress. 

\subsubsection{Value Guided Exploration (VGE)} This is the method described in this paper, combining sub-skill controllers and value-guided controller sampling.

\subsubsection{VGE + BC loss} This is a variant of our method that additionally uses behavior cloning with loss as in Eq. (\ref{eq:bcloss}).

\subsection{Results and discussion}

We evaluated the above experimental conditions individually for each of the sub-optimal controllers described in Sec \ref{sec:learnfg} (Fig ~\ref{sec:results}a-c). As we see, our method generally outperforms the baselines considered for each controller as proposed. 

The inability of baselines RL + BC loss and RL to learn is important to note. It underscores the key idea of our method that following actions sampled by the sub-skill controllers enables an otherwise very difficult exploration. 


Our evaluations also provide insights regarding the properties of the controllers that are critical for exploration. Fig \ref{fig:results} also demonstrates, that contact-switching behavior is critical for such exploration, as this controller enables learning finger-gaiting whenever it is used. 

While the in-grasp manipulation (IGM) controller is not critical for learning finger-gaiting, it is still beneficial as it improves training robustness. As seen from Table~\ref{tab:robust}, the success rate of learning to gait w.r.t varying seeds is markedly improved which may be attributable to the stability controller used as part of the in-grasp manipulation controller. 



Finally, we can also learn a single finger-gaiting policy for multiple objects with our method. In Fig~\ref{fig:multiobj}, we show that the finger-gaiting (FG) controller enables learning the finger-gaiting policy simultaneously for the sphere, cylinder, dodecahedron, icosahedron, and cube. The keyframes of the gaits achieved with this policy as visualized in Fig \ref{fig:gaits}.  

\begin{figure}[t]
\centering
\includegraphics[width=0.45\textwidth]{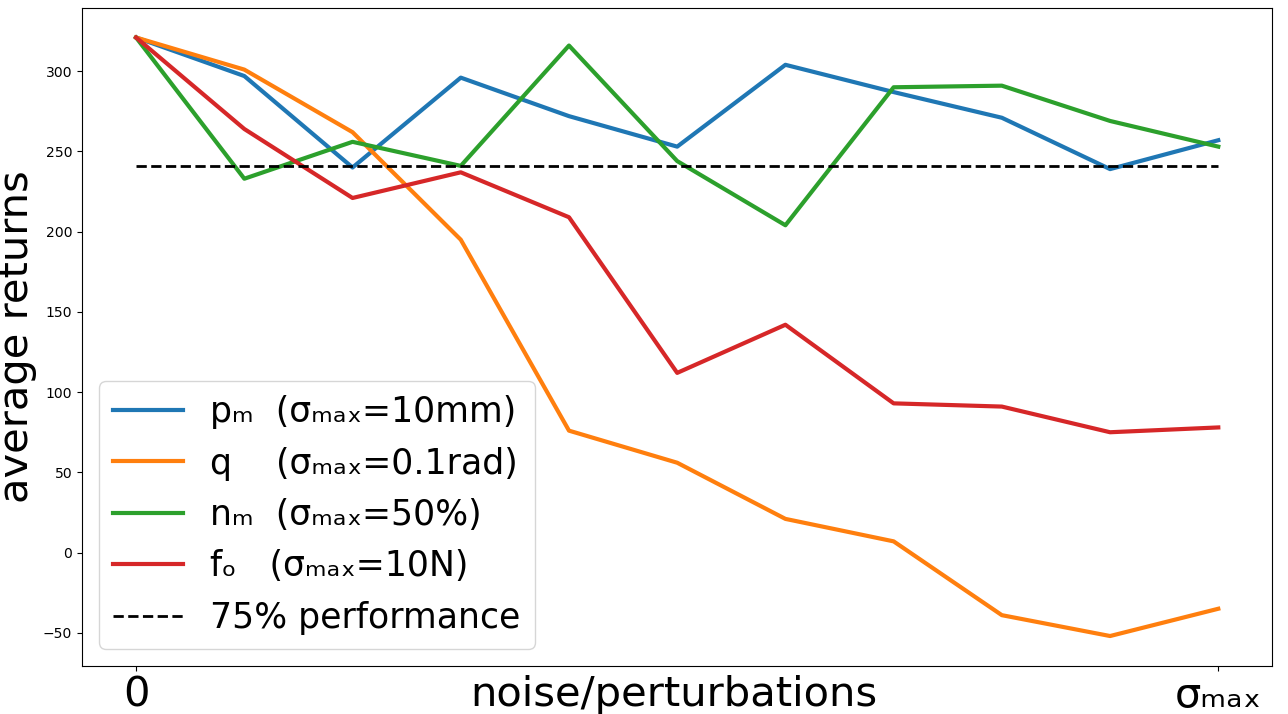}
\caption{Average episode returns of a policy trained on the cube with increasing perturbation forces and noise evaluated separately as shown. The policy sustains $0.1 rad$ of noise in joint positions $\boldsymbol{q}$, up to $50\%$  error in contact normal  $\boldsymbol{n}_m$ and a large error of $3mm$ in contact positions $\boldsymbol{p}_m$ -- all while suffering only  $25\%$  percent drop in average episode returns.}
\label{fig:perturbation}
\vspace{-5mm}
\end{figure}

\subsection{Sim-to-real transfer feasibility}
\label{sec:sim2real}
While we do not attempt sim-to-real transfer in this work, we believe that these policies can be transferred to the real hand using established methods from literature. As demonstrated by recent work \cite{OpenAI2018-bx, Chen2022-bm, Qi2022-wy, Khandate2023-gy}, domain randomization has been effective for transferring dexterous manipulation skills learned in simulation to the real hand. To investigate if our policies are suitable for such methods, we also tested their robustness to both perturbation forces on the object and noise in sensory feedback. As shown in Fig \ref{fig:perturbation}, our policies can sustain large perturbation forces equivalent to the full weight of the object, and are also robust to a high degree of sensor noise.  This characteristic leads us to believe that training with a curriculum of domain randomization will make transferring these robust policies to the real hand viable, and we hope to demonstrate this in future work.

\section{Conclusion}
Our work proposed the use of model-based controllers to assist exploration in reinforcement learning of dexterous manipulation tasks. To this end, we use off-policy RL and take advantage of the freedom with the construction of behavior policy to follow actions from the controller while collecting rollout trajectories to update the learning policy with such data to learn a successful policy.

We evaluated our method with the challenging dexterous manipulation task of learning finger-gaiting in-hand manipulation with only fingertip grasps. We designed simple controllers for key sub-skills in finger-gaiting - in-grasp manipulation and contact switching - and used these to enable learning effective finger-gaiting skills without any other additional means of improving exploration. Moreover, we show that our method benefits from model-based controllers even if the controllers are significantly sub-optimal.


More generally, we demonstrated a method to use domain expertise for learning dexterous manipulation tasks with improved sample efficiency. We believe our method of using sub-skill controllers for exploration is a promising approach for achieving dexterous skills with complex dynamics. While this work uses only model-based controllers, our method can also use learned sub-skill experts when available. It can also be extended to consider multiple experts. We hope to explore these promising directions in future work.


\addtolength{\textheight}{0cm}   
\clearpage
\printbibliography
\end{document}